\documentclass[a4paper, 10 pt, conference]{ieeeconf}      
\IEEEoverridecommandlockouts                              
\usepackage{cite}
\usepackage{amsmath,amssymb,amsfonts}
\usepackage{textcomp}
\usepackage[ruled,vlined,linesnumbered]{algorithm2e}
\usepackage[table]{xcolor}
\usepackage{pifont}
\usepackage{tikz}
\usepackage{multirow}
\usepackage{makecell}
\usepackage{color}

\newcommand{\cmark}{\ding{51}}%
\newcommand{\pmark}{\ding{115}}%
\newcommand{\xmark}{\ding{116}}%
\newcommand{\smark}{\ding{117}}%

\newcolumntype{C}[1]{>{\centering\let\newline\\\arraybackslash\hspace{0pt}}m{#1}}
\def\BibTeX{{\rm B\kern-.05em{\sc i\kern-.025em b}\kern-.08em
    T\kern-.1667em\lower.7ex\hbox{E}\kern-.125emX}}

\makeatletter
\def\endthebibliography{%
	\def\@noitemerr{\@latex@warning{Empty `thebibliography' environment}}%
	\endlist
}
\makeatother

\title{\LARGE \bf
On the Transferability of Knowledge among Vehicle Routing Problems by using Cellular Evolutionary Multitasking
}

\author{Eneko Osaba$^{1}$, Aritz D. Martinez$^{1}$, Jesus L. Lobo$^{1}$, Ibai La\~ na$^{1}$ and Javier Del Ser$^{1,2}$
\thanks{$^{1}$Eneko Osaba, Aritz D. Martinez, Jesus L. Lobo, Ibai La\~na and Javier Del Ser are with TECNALIA, Basque Research and Technology Alliance (BRTA), Derio, Spain. Contact email: eneko.osaba@tecnalia.com}%
\thanks{$^{2}$Javier Del Ser is with the University of the Basque Country (UPV/EHU), 48013 Bilbao, Bizkaia, Spain.}%
}

\begin{document}

\maketitle
\thispagestyle{empty}
\pagestyle{empty}

\begin{abstract}
Multitasking optimization is a recently introduced paradigm, focused on the simultaneous solving of multiple optimization problem instances (tasks). The goal of multitasking environments is to dynamically exploit existing complementarities and synergies among tasks, helping each other through the transfer of genetic material. More concretely, Evolutionary Multitasking (EM) regards to the resolution of multitasking scenarios using concepts inherited from Evolutionary Computation. EM approaches such as the well-known Multifactorial Evolutionary Algorithm (MFEA) are lately gaining a notable research momentum when facing with multiple optimization problems. This work is focused on the application of the recently proposed Multifactorial Cellular Genetic Algorithm (MFCGA) to the well-known Capacitated Vehicle Routing Problem (CVRP). In overall, 11 different multitasking setups have been built using 12 datasets. The contribution of this research is twofold. On the one hand, it is the first application of the MFCGA to the Vehicle Routing Problem family of problems. On the other hand, equally interesting is the second contribution, which is focused on the quantitative analysis of the positive genetic transferability among the problem instances. To do that, we provide an empirical demonstration of the synergies arisen between the different optimization tasks.
\end{abstract}

\section{Introduction} \label{sec:intro}

Transfer Optimization is a recently proposed paradigm, arisen with the idea of exploiting the knowledge learned from the optimization of one problem (or task), towards the facing of another related or unrelated task \cite{ong2016evolutionary}. This optimization paradigm has gained lot of attention within the community, chiefly because of two related reasons. On the one hand, the growing complexity of the optimization problems existing in the real-world. On the other hand, the needing of harnessing the lessons learned in previous related experiences.

Up to now, three different conceptualizations of the paradigm have been formulated \cite{gupta2017insights}: sequential transfer, multitasking and multiform optimization. The research introduced in this paper is focused on the second of these categories: multitasking. Specifically, multitasking is devoted to the simultaneous solving of distinct problem instances of equal importance by dynamically exploiting existing complementarities and synergies among tasks \cite{gupta2016genetic}. Particularly in this category, the correlation between tasks is of paramount importance for positively materialize the transfer of knowledge over the search \cite{gupta2015multifactorial}.

More specifically, we address multitasking through the perspective of Evolutionary Multitasking (EM, \cite{ong2016towards}) research trend, which embraces the main philosophy of Evolutionary Computation for simultaneously face several tasks at the same time. Going one step deeper, a particular EM conceptualization which has demonstrated a remarkable potential in last years is the coined as Multifactorial Optimization (MFO, \cite{gupta2015multifactorial}). This paradigm has been mainly materialized through the Multifactorial Evolutionary Algorithm (MFEA, \cite{gupta2015multifactorial}), which has focused most of the recent related literature \cite{wang2019evolutionary,xiao2019multifactorial,martinez2020simultaneously}.

In another vein, and thanks to the progressive advance of the related technology, transportation networks have become more complex along the years. This fact has led the mobility to be a crucial aspect for the society. On this regard, the necessity of an efficient transportation has become a cornerstone for both business companies and citizens. Within this problematic, route planning has arisen as an activity of paramount importance in Intelligent Transport Systems (ITS), easing the connection between Smart Cities and their inhabitants. 

Computationally speaking, routing problems are usually complex to solve. This fact, accompanied by the social interest that their solving entails, has made this kind of problems one of the most studied in Artificial Intelligence and Operations Research fields through last decades. Regarding their conceptualization, routing problems usually share some common characteristics that permit them to be modeled as optimization problems. Arguably, the most well-known examples of this kind of problems are the Traveling Salesman Problem (TSP, \cite{lin1965computer}) and the Vehicle Routing Problem (VRP, \cite{christofides1976vehicle}). Our research is focused on the Capacitated VRP (CVRP, \cite{ralphs2003capacitated}). Along the years, the CVRP and its variants has been studied throughout the perspective of different optimization paradigms, being the metaheuristic optimization the most representative one \cite{laporte2009fifty}.

In spite of the intense research activity conducted by the community around routing problems, they have scarcely addressed through the perspective of Transfer Optimization. Up to now, just two works have been published on this topic. The first one, published by Yuan et al. in 2016 \cite{yuan2016evolutionary}, is a conceptual research introducing the first adaptation of the MFEA to permutation-based combinatorial optimization problems. Among the benchmarking problems used for testing the feasibility of the proposed adaptation, the TSP is employed. On the second study, a similar adaptation of the MFEA is proposed by Zhou et al. \cite{zhou2016evolutionary}, in this case solely focused on the solving of the CVRP. Main novelty of this work is the split based decoding operator \cite{beasley1983route} used for translating the solutions of the unified search space to VRP solution space.

Despite being valuable and groundbreaking, these pioneer works are uniquely focused on the adaptation of the MFEA to discrete environments, and benchmarking solve of TSP and VRP. Bearing this background in mind, this present work aims to take a step further over the state of the art by elaborating on two research directions:
\begin{itemize}[leftmargin=*]
\item We adopt the recently proposed Multifactorial Cellular Genetic Algorithm (MFCGA, \cite{osaba2020MFCGA}) to the CVRP. MFCGA is a metaheuristic for MFO inspired by the MFEA and the foundations of Cellular Automata and Cellular Genetic Algorithms (cGAs \cite{manderick1989fine}). This philosophy helps the solver to control the mating process among different species (problems). Furthermore, the search strategy of the MFCGA enhances the detailed examination of synergies among the tasks being solved. This feature provides a better explainability interface for properly understand the positive genetic transfer arisen between tasks. In the current literature, the application of cGA concepts to EM scenarios is still scarce, supposing a significant contribution to the wider Transfer Optimization domain. Furthermore, it is the first time that the MFCGA is applied to VRP family of problems.

\item Aligned with some influential related works \cite{ong2016towards}, the performance of Transfer Optimization algorithms is directly related with the inter-tasks synergies of the problems involved. As claimed in previous studies \cite{zhou2018study}, the number of works delving on the analysis and measurement of similarities between optimization tasks and problems is really scant. Furthermore, in most of the cases, the deep assessment of the genetic transferability is a remarkably time-consuming and demanding task. For this reason, works such as \cite{gupta2016landscape} and \cite{da2017evolutionary}, which contribute to the field introducing insights in the measure of task relationship, suppose a remarkable contribution to the community. Having this issue in mind, we have conducted a deep analysis on the genetic transferability among CVRP instances, in order to establish the criteria that should be followed for assuring a profitable knowledge transfer among different CVRP cases.
\end{itemize}

This work is structured in the following way: in Section \ref{sec:back} we present a brief overview of background regarding EM, MFEA and cGAs. In \ref{sec:MFCGA} we introduce the main features of the MFCGA. The experimentation is detailed and discussed in Section \ref{sec:exp}. Section \ref{sec:conc} concludes the paper by drawing conclusions and outlining future research lines.

\section{Background} \label{sec:back}

This section is dedicated to introducing a short background about the two principal topics studied in this paper: EM/MFO (Section \ref{sec:back_EM}), and cGAs (Section \ref{sec:back_CGA}).

\subsection{Evolutionary Multitasking and Multifactorial Evolutionary Algorithm} \label{sec:back_EM}

EM has recently arisen as an efficient paradigm for facing Transfer Optimization situations. Until 2017, the concept of EM was only materialized through MFO concept \cite{da2017evolutionary}. This incipient stream of knowledge has attracted lot of attention in last years, and notably interesting approaches such as hybrid solvers \cite{xiao2019multifactorial}, multifactorial engines encompassing modern metaphors \cite{zheng2016multifactorial}, adaptations of classical algorithms \cite{feng2017empirical} or multipopulational methods \cite{song2019multitasking} have been introduced. Anyway, MFEA has led this incipient branch of the evolutionary computation field since its first formulation \cite{gupta2015multifactorial}.

MFO can be formulated as an environment in which $K$ optimization tasks should be simultaneously optimized. This environment is composed by multiple search spaces, as much as tasks to deal with. Supposing that all problems should be minimized, for the $k$-th task $T_k$ its objective function is conceived as $f_k : \Omega_k \rightarrow \mathbb{R}$, where $\Omega_k$ represents the solution space of $T_k$. Thus, the principal objective of MFO is to find a group of solutions $\{\mathbf{x}_1, \mathbf{x}_2,\dots,\mathbf{x}_K\}$ such that $\mathbf{x}_k = \arg \min_{\mathbf{x}\in\Omega_k} f_k(\mathbf{x})$. 

The main characteristic of MFO is that in spite of tackling $K$ isolated search processes, MFO seeks to find $\{\mathbf{x}_k\}_{k=1}^K$ by exploring a unified and unique search space $\Omega^\prime$. Accordingly, solutions $\mathbf{x}^\prime\in\Omega^\prime$ can represent a task-specific solution $\mathbf{x}_k$ for any of the $K$ tasks under optimization. Additionally, each individual $\mathbf{x}_p^\prime\in\Omega^\prime$ of the $P$-sized population is defined by the following four features:

\textit{Definition 1 (Factorial Cost)}: the factorial cost $\Psi_k^p$ of an individual $\mathbf{x}_p^\prime$ represents the value of the fitness function for a given task $T_k$. Each population member has a factorial cost list.

\textit{Definition 2 (Factorial Rank)}: the factorial rank $r_k^p$ of an population member $\mathbf{x}_p$ in a given task $T_k$ is the index of this member within the population sorted in ascending order of $\Psi_k^p$. Each population member has a factorial rank list.

\textit{Definition 3 (Scalar Fitness)}: the scalar fitness $\varphi^p$ of $\mathbf{x}_p^\prime$ is computed using the best factorial ranks over all the tasks in the following way: $\varphi^p = 1/ \left(\min_{k \in \{1...K\}}r_k^p\right)$.

\textit{Definition 4 (Skill Factor)}: The skill factor $\tau^p$ is the task in which an individual $\mathbf{x}_p^\prime$ performs best, namely, $\tau^p = \arg \min_{k\in\{1,\ldots,K\}} r_k^p$.

As mentioned above, MFEA is the most representative MFO method for solving EM environments \cite{gupta2015multifactorial}. This recently proposed algorithm sinks its roots on bio-cultural schemes of multifactorial inheritance. We represent in Algorithm \ref{alg:classicMFEA} the pseudo-code of the canonical variant of the MFEA. Additionally, we recommend the reading of \cite{gupta2015multifactorial} for deeper details on this solver. 

\begin{algorithm}[tb]
\small
	 \SetAlgoLined
		Randomly generate a population of $P$ individuals\;
		Evaluate each generated individual for the $K$ problems\;
		Calculate the skill factor ($\tau^p$) of each individual $\mathbf{x}_p^\prime$\;
		\Repeat{termination criterion not reached}{
		    Apply genetic operators on $P$ to get the offspring subpopulation $P_\ast$\;
		    Evaluate the generated offspring for the best task $\tau_\ast^p$ of their parent(s)\;
		    Combine $P$ and $P_\ast$ in intermediate population $Q$\;
		    Update the scalar fitness $\varphi_k^p$ and skill factor $\tau^p$ for each individual in $Q$\;
		    Build the next population by selecting the best $N$ individuals in $Q$ in terms of scalar fitness\;
		}
		Return the best individual in $P$ for each task $T_k$\;
 \caption{Pseudocode of the  MFEA}
 \label{alg:classicMFEA}
\end{algorithm}

For properly addressing the job of simultaneously optimize different tasks, MFEA has four keystone characteristics:

\textit{Unified search space}: the definition of a unified search space $\Omega^\prime$ able to represent feasible solutions for all the $k$ tasks under consideration is one of the main challenges when designing a MFEA. In this work, the permutation encoding is used as the unified representation for $\mathbf{x}_p^\prime$. Thus, considering $K$ VRP instances to be faced, and denoting the size of each instance $T_k$ (i.e. the number of \emph{clients}) as $D_k$, a possible solution $\mathbf{x}_p^\prime$ is represented as a permutation of the integer set $\{1,2,\ldots, D_{max}\}$, where $D_{max}=\max_{k\in\{1,\ldots,K\}} D_k$, namely, the maximum problem size among the $K$ tasks. Thus, each time an individual $\mathbf{x}_p^\prime$ has to be assessed on a task $T_k$ whose $D_k<D_{max}$, only values lower than $D_k$ are considered for producing solution $\mathbf{x}_k$ of $f_k(\cdot)$. Furthermore, zeros are dynamically introduced in the solution as control integers, in order to meet the capacity constraints of the problem.

\textit{Assortative mating}: this mating procedure establishes that individuals are prone to interact with other mates belonging to the same cultural background. Thus, as introduced in the pioneer work \cite{gupta2015multifactorial}, genetic operators used in MFEA enhance the mating among solutions with the same skill factor $\tau^p$. We highly recommend the analysis of that paper for deeper details on how this mechanism is implemented in MFEA.

\textit{Selective evaluation}: this feature is crucial to ensure the computational feasibility of the MFEA. Thus, selective evaluation implies that each newly generated solution is assessed only on a unique task. Concretely, each generated individual is measured in task $T_{\tau_\ast^p}$, where $\tau_\ast^p$ is the skill factor of its parent. The skill factor could also be randomly selected in cases in which the offspring has multiple parents. Furthermore, the factorial cost $\Psi_k^p$ is set to $\infty$ $\forall k\in\{1,\ldots,\tau_\ast^p-1,\tau_\ast^p+1,\ldots,K\}$.

\textit{Scalar fitness based selection}: this last feature regards the replacement strategy of the MFEA. In this case, this replacement is based on an elitist criterion, meaning that the best $P$ individuals in terms of scalar fitness $\sigma^p$ (considering both current population and the newly produced offspring) survive for the next generation.

\subsection{Cellular Genetic Algorithm} \label{sec:back_CGA}

Despite John von Neumann set out on the cellular automata journey in $1966$, they become fashionable with Conway's Game of Life years after \cite{games1970fantastic}. Anyway, the limelight really shone down on them when Stephen Wolfram presented his work \cite{wolfram2002new} in $2002$, where cellular automata definitely attracted the scientific attention. Since then, they have inspired many approaches that have solved real-world problems. One of the most successful is the cGA, which are a sub-type of the classical GAs in which the population is structured in a specific topology based on small-sized neighborhoods \cite{manderick1989fine}. 

In cGA, individuals only interact with their closest neighbors, which contributes to the exploration of the search space \cite{alba2004solving}. Regarding the exploitation, it is conducted within each neighborhood. Thus, the whole population in cGAs is structured over a grid, on which the aforementioned neighborhood relation is defined. Arguably, the most often used neighborhood structures are the NEWS (or von Neumann) and the C9 (or Moore). We recommend \cite{alba2009cellular} for additional information about cellular grid structures.

Furthermore, two categories of cGA can be distinguished depending on the update policy of the individuals. On the one hand, cGAs characterized by conducting all the replacements in parallel are coined as \textit{synchronous}. On the other hand, if solutions are update in a sequential way, cGAs are classified as \textit{asynchronous}. In our research, we have chosen this second scheme due to its ease adaptation to the newly generated genetic material \cite{alba2004solving}.

The literature is full of varied studies focused on cGAs and ITS. In \cite{pena2016multiobjective}, authors tackled the problem of vehicle scheduling in urban public transport systems considering the vehicle type and size, and using the well-known MOCell algorithm. In \cite{yu2015study} authors proposed an optimization model for addressing a stop-skipping problem with cGAs. A better understanding on traffic light scheduling has been recently provided by \cite{villagra2020better}. Regarding the field of VRP, in \cite{alba2004solving} authors present the first adaptation of the cGA to the CVRP. Authors of \cite{dorronsoro2007grid} proposed a grid-based hybrid cellular genetic algorithm for very large-scale instances of the CVRP. In \cite{kamkar2010cellular}, the authors focused on developing a cellular genetic algorithm for solving the VRP with time windows. The issue of complexity in CVRP using cGAs was tackled in \cite{niazy2012complexity}.

\section{Multifactorial Cellular Genetic Algorithm}\label{sec:MFCGA}

Influenced by the concepts of both MFEA and cGAs, the recently proposed MFCGA \cite{osaba2020MFCGA} acquires and reformulates the four MFEA pillars above introduced (unified representation, assortative mating, selective evaluation, and scalar fitness), using cGAs philosophy as inspiration. The main pseudocode of the MFCGA is represented in Algorithm \ref{alg:MFCGA}.

\begin{algorithm}[h]
\small
	 \SetAlgoLined
		Randomly generate an initial population $P$ of $N$ individuals\;
		Build the corresponding Moore Grids $MG_i = \{m_0,m_1,\dots,m_7\}$ for each individual $n_i$\;
		Evaluate each of the individual for all the $K$ optimization tasks\;
		Calculate the skill factor ($\tau$) of each individual\;
		
		\While{termination criterion not reached}{
		    \For{each individual $n_i$ in $N$}{
        		randomly select a neighbor $m_j$ from $MG_i$\; 
        		$n'_i$ $\gets$ crossover($n_i$,$m_j$)\;
        		$n''_i$ $\gets$ mutate($n_i$)\;
    		    Evaluate $n'_i$ and $n''_i$ for only the best task of $n_i$\;
    		    $n_i$ $\gets$ best($n_i$,$n'_i$,$n''_i$)\;
    		    Update scalar fitness ($\varphi$) and skill factor ($\tau$) of the new $n_i$\;
    		}
		}
		Return the best individual in $P$ for each task $T_k$\;
     \caption{Pseudocode of the MFCGA}
	 \label{alg:MFCGA}
\end{algorithm}

Similarly to as described for the MFEA, we describe now each crucial features of the MFGCA. First, as \textit{unified representation}, same encoding and procedure as in the case of the MFEA has been adopted.

The second important feature is the \textit{genetic operators mechanism}. This mechanism dictates the search procedure of the method, and it is partly inspired by the classical crossover and mutation functions of the cGAs. Thus, at each generation, every individual $n_i$ obligatorily goes through these two phases. This means that MFCGA does not use any kind of crossover or mutation probabilities. First of the newly generated individuals, $n_i'$, is the results of crossing $n_i$ with a neighbor $m_i$ chosen at random. This neighbor should be part of the Moore Grid neighborhood $MG_i$ of $i$, considering a neighborhood radius equal to 1. The second created solution, $n''_i$, is the result of a simple mutation of the basis individual $n_i$.

Once $n'_i$ and $n''_i$ are generated, and with the intention of assuring that MFGCA is computationally practical, the fitness of these two individuals are assessed following the same \textit{selective evaluation} used in MFEA. It is interesting to point here that $n'_i$ and $n''_i$ are measured on the task $T_\tau$, where $\tau$ is the skill factor of $n_i$. This mechanism entails a critical difference regarding MFEA, since it implies that individuals in MFGCA are committed to the optimization of the same task though the whole execution of the algorithm. In this respect, the equilibrium of the population is ensured since the first complete evaluation and sorting is conducted using the factorial rank and scalar factor. This way, the same number of solutions are dedicated to each of the optimization tasks.

Finally, as replacement criterion, a \textit{local improvement selection} mechanism has been implemented. This strategy establishes that the newly created $n'_i$ or $n''_i$ can only replace their parent $n_i$. More concretely, the only individual that survives to the next iteration is the best one among $n_i$, $n'_i$, and $n''_i$. 

\section{Experimentation and Results}\label{sec:exp}

This section is devoted to the detail and analysis of the conducted experimentation. Regarding the problem in which this paper is focused, the VRP, it is arguably one of the most studied problems in operations research and artificial intelligence fields. We refer interested readers to remarkable works such as \cite{kallehauge2008formulations} or \cite{kulkarni1985integer} for deepen the formulation and mathematical aspects of this classical problem. For readers interested on the VRP as a problem, we recommend the reading of surveys such as \cite{golden2008vehicle,osaba2020vehicle}. 

It is also interesting to highlight here that our main objective is not to find the optimal solutions of the used VRP instances. Instead, we use the VRP as benchmarking problem for measuring the performance of both MFEA and MFCGA. Additionally, a capital objective for us with this study is to properly analyze the genetic transfer arisen among the chosen datasets, and to conclude which features should meet the instances for previously ensure positive inter-task synergies. As foregrounded in previous published studies \cite{zhou2018study,gupta2016genetic}, this analysis supposes a remarkable contribution to the related literature.

For this purpose, 12 contrasted CVRP datasets have been employed in the present experimentation. All these instances are part of the well-known Augerat Benchmark \cite{augerat1995computational}. We list the instances used in Table \ref{tab:similarity}, accompanied with the summary of genetic complementarities present among them. In this sense, it is interesting to point that these datasests have been chosen due to their recognition by the related community, and because of the different levels of genetic similarities in their structure. In this case, we have measured these complementarities using as reference the amount of clients that the datasets share between them. Thus, our objective is to explore how these complementarities impact in the genetic exchange inherent to MFEA and MFCGA schemes. 

\begin{table*}[ht!]
    \centering
    \caption{Summary of genetic complementarities for all the datasets employed in the experimentation. Each cell represents the percentage of clients that dataset of the column has in common with the instance of the row.}
    \renewcommand{\arraystretch}{1.4}
    \resizebox{1.75\columnwidth}{!}{
        \begin{tabular}{c|c|c|c|c|c|c|c|c|c|c|c|c|}
            Instance & P-n16-k8 & P-n19-k2 & P-n20-k2 & P-n21-k2 & P-n22-k2 & P-n23-k8 & P-n50-k7 & P-n50-k8 & P-n55-k7 & P-n55-k15 & P-n60-k10 & P-n60-k15\\ \hline
            P-n16-k8 &  \cellcolor{gray!50} & 85\% & 88\% & 86\% & 84\% & 82\% & 3\% & 3\% & 3\% & 3\% & 2\% & 2\% \\ \hline
            P-n19-k2 &  100\% & \cellcolor{gray!50} & 97\% & 95\% & 92\% & 90\% & 3\% & 3\% & 3\% & 3\% & 2\% & 2\% \\ \hline
            P-n20-k2 &  100\% & 100\% & \cellcolor{gray!50} & 97\% & 95\% & 93\% & 3\% & 3\% & 3\% & 3\% & 2\% & 2\% \\ \hline
            P-n21-k2 &  100\% & 100\% & 100\% & \cellcolor{gray!50} & 97\% & 95\% & 3\% & 3\% & 3\% & 3\% & 2\% & 2\% \\ \hline
            P-n22-k2 &  100\% & 100\% & 100\% & 100\% & \cellcolor{gray!50} & 97\% & 3\% & 3\% & 3\% & 3\% & 2\% & 2\% \\ \hline
            P-n23-k8 &  100\% & 100\% & 100\% & 100\% & 100\% & \cellcolor{gray!50} & 3\% & 3\% & 3\% & 3\% & 2\% & 2\% \\ \hline
            P-n50-k7 &  3\% & 3\% & 3\% & 3\% & 3\% & 3\% & \cellcolor{gray!50} & 100\% & 95\% & 95\% & 90\% & 90\% \\ \hline
            P-n50-k8 &  3\% & 3\% & 3\% & 3\% & 3\% & 3\% & 100\% & \cellcolor{gray!50} & 95\% & 95\% & 90\% & 90\% \\ \hline
            P-n55-k7 &  3\% & 3\% & 3\% & 3\% & 3\% & 3\% & 100\% & 100\% & \cellcolor{gray!50} & 100\% & 95\% & 95\\ \hline
            P-n55-k15 &  3\% & 3\% & 3\% & 3\% & 3\% & 3\% & 100\% & 100\% & 100\% & \cellcolor{gray!50} & 95\% & 95\\ \hline
            P-n60-k10 &  2\% & 2\% & 2\% & 2\% & 2\% & 2\% & 100\% & 100\% & 100\% & 100\% & \cellcolor{gray!50} & 100\% \\ \hline
            P-n60-k15 &  2\% & 2\% & 2\% & 2\% & 2\% & 2\% & 100\% & 100\% & 100\% & 100\% & 100\% & \cellcolor{gray!50}\\ \hline
        \end{tabular}
    }
    \vspace{2mm}
    \label{tab:similarity}
\end{table*}

\subsection{Experimental Setup}\label{sec:exp_setup}

For properly assess the performance of the MFCGA, we have compared its results with the ones obtained by the canonical version of the MFEA, which is the current baseline in the related literature. For adequately parameterize both metaheuristics, studies focused on cGAs and MFEA have also been used as inspiration \cite{alba2004solving,yuan2016evolutionary,zhou2016evolutionary}. For the sake of replicability, the parameterizations employed for MFCGA and MFEA are shown in Table \ref{tab:Parametrization}. Moreover, all individuals are randomly generated, and each metaheuristic uses a fixed number of 50k objective function evaluations as termination criterion.

\begin{table}[tb]
	\centering
	\caption{Parameterization of MFCGA and MFEA}
	\resizebox{0.85\columnwidth}{!}{
		\begin{tabular}{ l || l | l |}
			Parameter & MFCGA & MFEA\\
			\hline
			Population size & 200 & 200 \\ \hline
			Crossover Function & Order Crossover & Order Crossover \\ \hline
			Mutation Function & 2-opt & 2-opt \\ \hline
			Crossover Prob. & 1.0 & 0.9 \\ \hline
			Mutation Prob. & 1.0 & 0.1 \\ \hline
			Type of grid & Moore & \\ \hline
		\end{tabular}
	}
	\vspace{2mm}
	\label{tab:Parametrization}
\end{table}

On another level, 11 different test cases have been generated using the 12 datasets above mentioned. Each of these multitasking scenarios involves that MFEA and MFCGA should face the resolutions of all the tasks assigned to that configurations. Summarizing, among the test cases built, 4 of them are composed by four instances, 6 contemplate the solving of six datasets, and the last test case is comprised by all the 12 tasks. Table \ref{tab:testCases} shows all the considered configurations. The reasons of constructing these 11 tests cases is twofold. First, we try to enhance the heterogeneity and variety of the configurations, using each instance in the same amount of configurations. The second reason regards the effort made to exploit the complementarities represented in Table \ref{tab:similarity}.

\begin{table*}[ht!]
    \centering
    \caption{Summary of the 11 tests cases generated}
    \renewcommand{\arraystretch}{1.4}
    \resizebox{1.5\columnwidth}{!}{
        \begin{tabular}{c|c|c|c|c|c|c|c|c|c|c|c|c|}
            Test Case &  P-n16-k8 & P-n19-k2 & P-n20-k2 & P-n21-k2 & P-n22-k2 & P-n23-k8 & P-n50-k7 & P-n50-k8 & P-n55-k7 & P-n55-k15 & P-n60-k10 & P-n60-k15\\ \hline
            TC\_4\_1 &  \centering \cmark &  \centering \cmark &  \centering \cmark &  \centering \cmark & -- & -- & -- & -- & -- & -- & -- & -- \\ \hline
            TC\_4\_2 & -- &  -- &  \centering \cmark &  \centering \cmark & \centering \cmark & \centering \cmark & -- & -- & -- & -- & -- & -- \\ \hline
            TC\_4\_3 & -- & -- & -- & -- & -- & -- & \centering \cmark &  \centering \cmark &  \centering \cmark &  \centering \cmark & -- & -- \\ \hline
            TC\_4\_4 & -- & -- & -- & -- & -- & -- & -- & -- & \centering \cmark &  \centering \cmark & \centering \cmark &  {\centering \cmark} \\ \hline
            TC\_6\_1 & \centering \cmark & \centering \cmark & \centering \cmark & -- & -- & -- & \centering \cmark & \centering \cmark & \centering \cmark & -- & -- & --\\ \hline
            TC\_6\_2 & -- & -- & -- & \centering \cmark & \centering \cmark & \centering \cmark & -- & -- & -- &  \centering \cmark & \centering \cmark &  {\centering \cmark} \\ \hline
            TC\_6\_3 & \centering \cmark & \centering \cmark & \centering \cmark & -- & -- & -- & -- & -- & -- & \centering \cmark & \centering \cmark & {\centering \cmark} \\ \hline
            TC\_6\_4 & -- & -- & -- & \centering \cmark & \centering \cmark & \centering \cmark &  \centering \cmark & \centering \cmark &  {\centering \cmark} & -- & -- & -- \\ \hline
            TC\_6\_5 & \centering \cmark & \centering \cmark & \centering \cmark & \centering \cmark & \centering \cmark & \centering \cmark & -- & -- & -- & -- & -- & -- \\ \hline
            TC\_6\_6 & -- & -- & -- & -- & -- & -- & \centering \cmark & \centering \cmark & \centering \cmark &  \centering \cmark & \centering \cmark &  {\centering \cmark}\\ \hline
            TC\_12 & \centering \cmark & \centering \cmark & \centering \cmark & \centering \cmark & \centering \cmark & \centering \cmark & \centering \cmark & \centering \cmark & \centering \cmark &  \centering \cmark & \centering \cmark &  {\centering \cmark} \\ \hline
        \end{tabular}
    }
    \vspace{2mm}
    \label{tab:testCases}
\end{table*}

\subsection{Results and discussion}\label{sec:exp_res}

\begin{table}[ht!]
    \centering
    \caption{Comparison of the results for the 15 tests cases built for the experimentation. \pmark \hspace{0.02cm} means MFCGA outperforms MFEA. \smark \hspace{0.02cm} depicts both alternatives have performed similar. \xmark \hspace{0.02cm} means MFEA performs better.}
    \renewcommand{\arraystretch}{1.4}
    \resizebox{0.85\columnwidth}{!}{
        \begin{tabular}{c|C{2cm}|c|C{2cm}|}
            Test Case & Comparison &  Test Case & Comparison \\ \hline
            TC\_4\_1 & \smark \pmark \pmark \pmark & TC\_6\_3 & \smark \pmark \pmark \pmark \pmark \pmark \\ \hline
            TC\_4\_2 & \pmark \pmark \pmark \pmark & TC\_6\_4 & \pmark \pmark \pmark \pmark \pmark \pmark\\ \hline
            TC\_4\_3 & \pmark \pmark \pmark \pmark & TC\_6\_5 & \smark \pmark \pmark \pmark \pmark \pmark\\ \hline
            TC\_4\_4 & \xmark \pmark \pmark \pmark & TC\_6\_6 & \pmark \pmark \pmark \pmark \pmark \pmark\\ \hline
            TC\_6\_1 & \smark \pmark \xmark \pmark \pmark \pmark & \multirow{2}{*}{TC\_12} & \multirow{2}{*}{\thead{\smark \pmark \pmark \pmark \pmark \pmark \\ \pmark \pmark \pmark \pmark \pmark \pmark}}\\ \cline{0-1}
            TC\_6\_2 & \pmark \pmark \pmark \pmark \pmark \pmark & & \\ \hline
        \end{tabular}
    }
    \vspace{2mm}
    \label{tab:testCasesSummary}
\end{table}

We summarize in Table \ref{tab:testCasesSummary} the comparison of the results reached by MFEA and MFCGA. For obtaining statistically significant findings, each tests case has been run 20 times. For accommodating to the length limitation of the paper, we graphically depict the comparison in the results using two different signs, instead of showing all the average outcomes. Concretely, we represent as \pmark \hspace{0.02cm} when MFCGA outperforms MFEA in terms of fitness average, and as \xmark \hspace{0.02cm} otherwise. Furthermore, \smark \hspace{0.02cm} denotes that both algorithms reached same results. For understanding the table, using $TC\_4\_1$ as example and the instance order shown in \ref{tab:testCases}, we can see how MFCGA and MFEA obtain similar results for \textit{P-n16-k8} dataset, while in the case of \textit{P-n19-k2}, \textit{P-n20-k2} and \textit{P-n21-k2} MFCGA ourperforms MFEA. With this consideration, we can clearly observe how MFCGA shows a better efficiency of facing the build 11 multitasking configurations, being outperformed by MFEA is only two cases. It is especially remarkable the performance shown in the last test case $TC\_12$, in which MFCGA obtains better results in all but one dataset.

Aiming to enhance the completeness of this study, we depict in Table \ref{tab:bestSolutions} the outcomes obtained by MFCGA and MFEA in all the considered datasets. These results have been obtained after the 20 runs of the \textit{TC\_12}. Results represented in that table support the findings drawn in the previous Table \ref{tab:testCasesSummary}. MFGCA outperforms its counterpart in all datasets in terms of average results, except in the case of \textit{P-n16-k8} in which both solvers reach the optima in all the 20 runs. Regarding the best solution found, MFCGA also emerges victorious. Lastly, and despite not being the goal of this work, it can be observed how the deviation regarding known optimal results and the average outcomes obtained by the MFCGA ranges between 0'0\% and 5'8\%. This performance clearly allows us to prudently affirm that MFCGA is a promising method for solving CVRP datasets. 

Additionally, we show the outcomes obtained by the Wilcoxon Rank-Sum test in the last row of the Table \ref{tab:bestSolutions}. We represent graphically the results got. Specifically, \pmark \hspace{0.02cm} depicts that MFCGA significantly outperforms MFEA, while \smark \hspace{0.02cm} means that there is not enough evidence to state that the difference is statistically remarkable. The confidence interval has been set in $95$\%. As a summary, Wilcoxon Rank-Sum test confirms that the MFCGA approach is statistically better in 9 out of 12 of the instances. This fact supports the fact that MFCGA is a promising method for solving EM environments based on the CVRP.

\begin{table}[ht!]
    \centering
    \caption{Results obtained by MFCGA and MFEA for the 12 dataset that compose test case $TC\_12$, and graphical results of the Wilcoxon Rank-Sum test.}
    \renewcommand{\arraystretch}{1.4}
    \resizebox{0.90\columnwidth}{!}{
        \begin{tabular}{c|c|c|c|c|c|c|}
            & P-n16-k8 & P-n19-k2 & P-n20-k2 & P-n21-k2 & P-n22-k2 & P-n23-k8 \\ \hline
            MFCGA & 450.0 & 212.0 & 216.0 & 211.0 & 216.0 & 530.9 \\
            & 450.0 & 212.0 & 216.0 & 211.0 & 216.0 & 529.0 \\
            & 0.0 & 0.0 & 0.0 & 0.0 & 0.0 & 2.43 \\ \hline
            
            MFEA & 450.0 & 219.6 & 224.6 & 211.5 & 216.6 & 538.4 \\
            & 450.0 & 212.0 & 217.0 & 211.0 & 216.0 & 529.0 \\
            & 0.0 & 6.24 & 5.18 & 1.02 & 0.91 & 4.05 \\ \hline\hline

            Optima & 450 & 212 & 216 & 211 & 216 & 529 \\ \hline \hline
            
			&\multicolumn{6}{c|}{Wilcoxon Rank-Sum test}\\
			\hline
			& \smark & \pmark & \pmark & \smark & \smark & \pmark \\ \hline\hline
            
            & P-n50-k7 & P-n50-k8 & P-n55-k7 & P-n55-k15 & P-n60-k10 & P-n60-k15 \\ \hline
            
            MFCGA & 571.9 & 679.8 & 604.0 & 982.7 & 781.3 & 1024.9 \\
            & 578.0 & 656.0 & 589.0 & 962.0 & 755.0 & 980.0 \\
            & 6.49 & 10.36 & 8.66 & 10.91 & 15.12 & 14.68 \\ \hline
            
            MFEA & 612.5 & 683.5 & 641.3 & 1021.1 & 812.3 & 1089.4 \\
            & 581.0 & 660.0 & 606.0 & 981.0 & 785.0 & 1008.0 \\
            & 20.82 & 17.75 & 32.47 & 30.07 & 35.25 & 23.42 \\ \hline\hline

            Optima & 554 & 629 & 568 & 945 & 744 & 968 \\ \hline\hline
            
			&\multicolumn{6}{c|}{Wilcoxon Rank-Sum test}\\
			\hline
			& \pmark & \pmark & \pmark & \pmark & \pmark & \pmark \\ \hline
            \end{tabular}
    }
    \vspace{2mm}
    \label{tab:bestSolutions}
\end{table}

\subsection{Analysis of the genetic transfer}\label{sec:exp_gen}

This section is devoted to delving in the genetic transfer cropped up across the considered 12 CVRP datasets. We focus this analysis in the activity arisen through the application of the MFCGA. This metaheuristic is especially interesting for these purposes because of the replacement strategy employed. In MFCGA an individual $n_i$ is replaced if, and only if, any of the individuals created through the mutation ($n''_i$) and crossover ($n'_i$) mechanisms outperform $n_i$ regarding its best performing task. This way, if $n'_i$ substitutes $n_i$, we can firmly asset that a positive genetic material transfer has occurred among $n_i$ and $m_j$ (we refer to Algorithm \ref{alg:MFCGA} in Section \ref{sec:MFCGA} for notation details) and that an exchange of genetic material has been explicitly produced. 

This being explained, main objectives with the study depicted in this section are threefold: to discover the synergies inherent to the used tasks, to objectively glimpse the positive inter-task interactions arisen, and to scrutinize the transfer of knowledge emerged. We have chosen the test case \textit{TC\_12} for this analysis because of being the only case in which the whole deemed CVRP instances are simultaneously optimized. In the context of the problem at hand, the positive genetic transfer takes place through the direct replication of an extract of the solution $m_j$ into $n'_i$.

\begin{figure}[t]
    \centering
    \includegraphics[width=1.0\hsize]{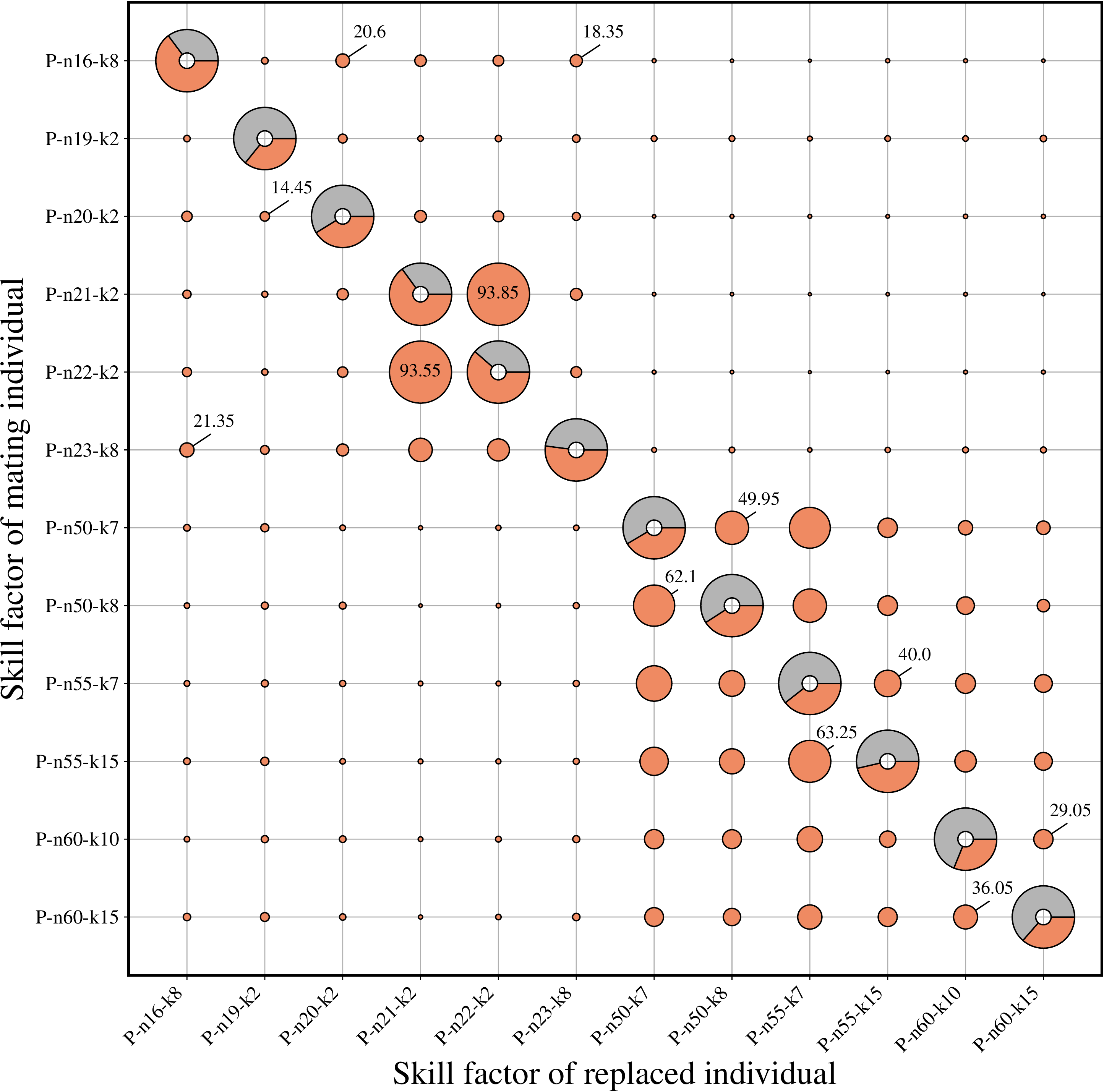}
    \caption{Intensities of the genetic transfer between the considered tasks.}
    \label{fig:matrix_influence}
\end{figure}

We summarize in Figure \ref{fig:matrix_influence} the intensities of positive inter-task interactions between the chosen tasks. In this graphic, the thickness of each orange circle represents the average number of times per run in which a solution with the skill factor of the column has positively transferred some of its genetic material to an individual which performs best in the task of the row. Furthermore, diagonal circles depict the combination of intra-task interactions (gray portion, representing the positive transfer of knowledge between solutions with the same skill factor) and the sum of all arisen inter-task exchanges (orange portion).

After thoroughly examining this figure, two valuable findings can be drawn. The first of these findings is the confirmation of the existence of intra-task synergy between the chosen dataset, materialized in the significant positive genetic transfer arisen between the tasks. This situation is specially representative in tasks pairs such as \{\textit{P-n21-k2---P-n22-k2}\}, \{\textit{P-n50-k7---P-n55-k8}\} and \{\textit{P-n50-k7---P-n55-k7}\}. Furthermore, it can be clearly glimpsed in Figure \ref{fig:matrix_influence} how the selected instances are divided into two different families. On the one hand, we have the group comprised by \{\textit{P-n16-k8,P-n19-k2,P-n20-k2,P-n21-k2,P-n22-k2,P-n23-k8}\}. On the other hand, we can distinguish the set composed by \{\textit{P-n50-k7,P-n50-k8,P-n55-k7,P-n55-k15,P-n60-k10,P-n60-k15}\}. In this way, we can clearly see how the intra-task activity inside these groups is intense, while the positive genetic sharing is almost nonexistent when a task is mated with an individual specialized in an instance which is outside its family. 

The second conclusion that can be extracted is that, despite the intense inter-task activity between tasks of the same family, these relations differ depending on the pair of datasets that are being mated. An insightful reader may correctly ask why the intensities between, for example, pairs \{\textit{P-n21-k2---P-n22-k2}\} and \{\textit{P-n21-k2---P-n23-k8}\} are so different. Analyzing the correlation in the landscapes of these tasks, we can confirm that the so-called partial domain overlap exists \cite{gupta2017insights}, having a subset of features that are common in the three datasets. In fact, the genetic complementarity between \textit{P-n21-k2} and \textit{P-n22-k2} is 97\% according to Table \ref{tab:similarity}, and the one regarding \textit{P-n21-k2} and \textit{P-n23-k8} is 95\%.

To properly understand these contradictions, a much deeper analysis should be conducted with the structure of the whole considered 12 CVRP datasets. Recently published works such as \cite{zhou2018study} and \cite{da2017evolutionary} suggest the use of the correlation among the best-known solutions as an appropriate approach for understanding the inter-task genetic transfer activity. Embracing this consideration, we have used this approach in our analysis.

Therefore, we summarize the correlation in the optimal solutions of the 12 CVRP datasets in Table \ref{tab:similarityBestSolutions}. Additionally, in order to enhance the visibility of this analysis, we have coloured in orange those cells corresponding to the pairs of tasks that have shown a noteworthy positive inter-task genetic material transfer. In fact, with the intention of being more specific, the more intense the inter-task activity, the more intense the orange used for coloring the cell. We have conducted this in an attempt of turning Table \ref{tab:similarityBestSolutions} those conclusions drawn Figure \ref{fig:matrix_influence}.

\begin{table*}[ht!]
    \centering
    \caption{Genetic complementarities in the optimal solutions of the datasets employed in the experimentation.}
    \renewcommand{\arraystretch}{1.4}
    \resizebox{1.75\columnwidth}{!}{
        \begin{tabular}{c|c|c|c|c|c|c|c|c|c|c|c|c|}
            Instance & P-n16-k8 & P-n19-k2 & P-n20-k2 & P-n21-k2 & P-n22-k2 & P-n23-k8 & P-n50-k7 & P-n50-k8 & P-n55-k7 & P-n55-k15 & P-n60-k10 & P-n60-k15\\ \hline
            P-n16-k8 &  \cellcolor{gray!50} & \cellcolor{orange!13}13\% & \cellcolor{orange!50}33\% & \cellcolor{orange!28}26\% & \cellcolor{orange!28}40\% & \cellcolor{orange!50}53\% & 0\% & 0\% & 0\% & 0\% & 0\% & 0\% \\ \hline
            P-n19-k2 &  \cellcolor{orange!13}11\% & \cellcolor{gray!50} & \cellcolor{orange!28}77\% & \cellcolor{orange!13}33\% & \cellcolor{orange!13}33\% & \cellcolor{orange!13}16\% & 0\% & 0\% & 0\% & 0\% & 0\% & 0\%\\ \hline
            P-n20-k2 &  \cellcolor{orange!28}26\% & \cellcolor{orange!28}73\% & \cellcolor{gray!50} & \cellcolor{orange!28}63\% & \cellcolor{orange!28}63\% & \cellcolor{orange!13}26\% & 0\% & 0\% & 0\% & 0\% & 0\% & 0\% \\ \hline
            P-n21-k2 &  \cellcolor{orange!13}20\% & \cellcolor{orange!13}30\% & \cellcolor{orange!28}60\% & \cellcolor{gray!50} & \cellcolor{orange!80}\textbf{80}\% & \cellcolor{orange!28}35\% & 0\% & 0\% & 0\% & 0\% & 0\% & 0\% \\ \hline
            P-n22-k2 &  \cellcolor{orange!13}28\% & \cellcolor{orange!13}28\% & \cellcolor{orange!28}57\% & \cellcolor{orange!80}\textbf{76}\% & \cellcolor{gray!50} & \cellcolor{orange!28}28\% & 0\% & 0\% & 0\% & 0\% & 0\% & 0\% \\ \hline
            P-n23-k8 &  \cellcolor{orange!28}36\% & \cellcolor{orange!13}13\% & \cellcolor{orange!13}22\% & \cellcolor{orange!50}31\% & \cellcolor{orange!50}27\% & \cellcolor{gray!50} & 0\% & 0\% & 0\% & 0\% & 0\% & 0\% \\ \hline
            P-n50-k7 &  0\% & 0\% & 0\% & 0\% & 0\% & 0\% & \cellcolor{gray!50} & \cellcolor{orange!35}51\% & \cellcolor{orange!60}59\% & \cellcolor{orange!28}26\% & \cellcolor{orange!13}34\% & \cellcolor{orange!13}34\% \\ \hline
            P-n50-k8 &  0\% & 0\% & 0\% & 0\% & 0\% & 0\% & \cellcolor{orange!60}51\% & \cellcolor{gray!50} & \cellcolor{orange!50}46\% & \cellcolor{orange!35}32\% & \cellcolor{orange!35}30\% & \cellcolor{orange!20}38\% \\ \hline
            P-n55-k7 &  0\% & 0\% & 0\% & 0\% & 0\% & 0\% & \cellcolor{orange!50}53\% & \cellcolor{orange!35}42\% & \cellcolor{gray!50} & \cellcolor{orange!35}31\% & \cellcolor{orange!20}37\% & \cellcolor{orange!20}35\%\\ \hline
            P-n55-k15 &  0\% & 0\% & 0\% & 0\% & 0\% & 0\% & \cellcolor{orange!30}24\% & \cellcolor{orange!30}29\% & \cellcolor{orange!60}31\% & \cellcolor{gray!50} & \cellcolor{orange!20}24\% & \cellcolor{orange!20}44\%\\ \hline
            P-n60-k10 &  0\% & 0\% & 0\% & 0\% & 0\% & 0\% & \cellcolor{orange!20}28\% & \cellcolor{orange!20}25\% & \cellcolor{orange!35}33\% & \cellcolor{orange!20}22\% & \cellcolor{gray!50} & \cellcolor{orange!20}40\% \\ \hline
            P-n60-k15 &  0\% & 0\% & 0\% & 0\% & 0\% & 0\% & \cellcolor{orange!20}28\% & \cellcolor{orange!20}32\% & \cellcolor{orange!35}32\% & \cellcolor{orange!20}40\% & \cellcolor{orange!35}40\% & \cellcolor{gray!50}\\ \hline
        \end{tabular}
    }
    \vspace{2mm}
    \label{tab:similarityBestSolutions}
\end{table*}

Several interesting trends can be seen in Table \ref{tab:similarityBestSolutions}. First of all, it can be seen how the pairs with the higher positive transfer activity present a significant overlap in their optimal solutions. This affirmation is visible in cases such as \{\textit{P-n21-k2---P-n22-k2}\}, \{\textit{P-n50-k7---P-n55-k7}\} or \{\textit{P-n50-k7---P-n50-k8}\}. On the contrary, it can be observed how pairs of tasks with a lower level of synergy in their optimal solution present a dimmer intensity on their positive genetic transfer. \{\textit{P-n16-k8---P-n19-k2}\}, \{\textit{P-n23-k8---P-n19-k2}\} or \{\textit{P-n19-k2---P-n23-k8}\} are examples that support this claim. Performing a complete analysis of the table, it can be seen how this trend is representative in the majority cases. Obviously, there are some examples in which this trend is not strictly adhered to, due to the randomness of the metaheuristics and the random generation of the Moore grids. In any case, the coherence is maintained in all cases, in the sense that pairs with a high degree of intersection are more likely to have an intense positive intra-task knowledge transfer. It should be pointed that we use the definition of \textit{intersection} provided in\cite{da2017evolutionary}. Thus, we can say that two tasks are intersected if \textit{the global optima of the two tasks are identical in the unified search space with respect to a subset of variables only, and different with respect to the remaining variables}.

This deeper study has led us to the last finding of the research presented on this paper: for VRP problems, the positive material transfer among different tasks is strictly related to the degree of intersection in their best solution. More concretely, we have shown with our study that intersection degrees greater that 11\% are enough for ensuring a minimum positive activity. Furthermore, tasks with greater degrees of similarity are prone to present a more intense knowledge transfer. Also, this fact led us to confirm that the sole complementarity in the structure of the VRP dataset is irrelevant for the genetic transfer.

\section{Conclusions and Future Work}\label{sec:conc}

This paper has been devoted to the adaptation of the Multifactorial Cellular Genetic Algorithm for solving the Capacitated Vehicle Routing Problem. First, we have presented the method, which sinks its roots in the main concepts of the cellular Genetic Algorithms and the well-known Multifactorial Evolutionary Algorithm. For the experimentation, 12 recognized instances of the CVRP, part of the Augerat benchmark, have been employed for generating 11 multitasking environments. For measuring the performance of the MFCGA, we have used the canonical MFEA as control algorithm. Obtained results have confirmed that the MFCGA is a promising solver for solving CVRP multitasking scenarios.

Even more interesting is the inter-task genetic transfer analysis carried out with the 12 datasets, in an effort to uncover the synergies that should be exist between to CVRP instances for maximizing the performance of the multitasking solving schemes. Our principal finding on this respect is that the positive genetic transfer is prone to occur when tasks share a minimum of their genetic structure on their best solution.

As future work, we have planned several lines to continue the preliminary research presented in this paper. First, we intend to continue exploring the CVRP as a problem, using larger instances and larger test cases. These additional tests will also contribute to measure the scalability of the proposed metaheuristic. We have also planned the using of additional mechanisms on the MFCGA, such as heuristic local search methods or alternative survivor strategies. We have established as long term future work the application of the MFCGA to other optimization fields, aiming to discover further intra-task genetic transfer synergies on valuable combinatorial optimization tasks.

\section*{Acknowledgment}

Eneko Osaba, Aritz D. Martinez, Jesus L. Lobo, Ibai Laña and Javier Del Ser would like to thank the Basque Government for its funding support through the EMAITEK and ELKARTEK programs. Javier Del Ser receives funding support from the Consolidated Research Group MATHMODE (IT1294-19) granted by the Department of Education of the Basque Government.

\bibliographystyle{IEEEtran}
\bibliography{IEEEexample}

\end{document}